\documentclass{article}
\usepackage{amssymb}
\usepackage{amsmath}
\usepackage{graphicx}
\usepackage{algorithm}
\usepackage{algorithmic}
\usepackage{booktabs}
\usepackage{makecell}
\usepackage{wrapfig}
\usepackage{multirow} 
\usepackage{caption}
\usepackage{hyperref}
\usepackage[utf8]{inputenc}

\usepackage[final]{corl_2025} 

\title{Constrained Style Learning from Imperfect Demonstrations under Task Optimality}

\author{
  Kehan Wen\\
  ETH Zurich\\
  \texttt{kehwen@ethz.ch} \\
  \And
  Chenhao Li \\
  ETH AI Center\\
 \texttt{chenhao.li@ai.ethz.ch} \\
  \And
  Junzhe He \\
  ETH Zurich\\
  \texttt{junzhe@ethz.ch} \\
  \AND
  Marco Hutter \\
  ETH Zurich\\
  \texttt{mahutter@ethz.ch} \\
}

\begin{document}
\maketitle


\begin{abstract}
Learning from demonstration has proven effective in robotics for acquiring natural behaviors, such as natural motions and lifelike agility, particularly when explicitly defining style-oriented reward functions is challenging. Synthesizing stylistic motions for real-world tasks usually requires balancing task performance and imitation quality. Existing methods generally depend on expert demonstrations closely aligned with task objectives. However, practical demonstrations are often incomplete or unrealistic, causing current methods to boost style at the expense of task performance. To address this issue, we propose formulating the problem as a constrained Markov Decision Process (CMDP). Specifically, we optimize a style-imitation objective with constraints to maintain near-optimal task performance. We introduce an adaptively adjustable Lagrangian multiplier to guide the agent to imitate demonstrations selectively, capturing stylistic nuances without compromising task performance. We validate our approach across multiple robotic platforms and tasks, demonstrating both robust task performance and high-fidelity style learning. On ANYmal-D hardware we show a $14.5 \%$ drop in mechanical energy and a more agile gait pattern, showcasing real-world benefits.
\end{abstract}

\keywords{Constrained Markov Decision Process, Imitation Learning, Legged Robots} 

\section{Introduction}

Reinforcement learning (RL)~\citep{sutton1999reinforcement} has demonstrated significant potential to achieve robust and adaptive control of legged robots, due to its capability to handle uncertainties in real-world applications~\citep{tan2018sim, lee2020learning, hwangbo2019learning, miki2022learning, kumar2021rma, margolis2024rapid, hoeller2024anymal, jenelten2024dtc, xie2020learning, siekmann2021blind, duan2024learning}. Despite this success, integrating fine-grained stylistic behaviors, such as agile locomotion or expressive motions, remains challenging. Although RL excels at optimizing clear, task-specific goals (e.g., velocity tracking, goal-reaching), crafting reward functions for nuanced, high-dimensional stylistic behaviors is inherently difficult~\citep{lee2020learning, margolis2024rapid, siekmann2021sim}.

Learning from Demonstration (LfD)~\citep{schaal1996learning} has emerged as a powerful technique that embeds stylistic imitation in robotic behaviors. Traditional LfD techniques include motion-clip tracking~\citep{peng2018deepmimic} and adversarial imitation learning~\citep{peng2021amp, li2023learning, li2023versatile}. However, obtaining high-quality reference motions that perfectly align with task objectives is often labor intensive (e.g., motion capture). Most practical demonstrations contain imperfections, such as incomplete data collection or unrealistic motions from biological counterparts with different morphologies~\citep{li2024fld}. For example, human motion data collected on flat terrain may poorly generalize to tasks involving complex, uneven terrains, causing suboptimal task performance. Strict adherence to these demonstrations often leads to suboptimal outcomes, emphasizing the critical trade-off between stylistic fidelity and task effectiveness.

A common approach is to manually tune fixed weights or craft a customized curriculum, a tedious process that provides \textbf{no guarantee} of near-optimal task reward. Inspired by work from safe reinforcement learning domain—where constrained Markov decision processes (CMDPs) are commonly used to enforce safety constraints~\citep{achiam2017constrained, garcia2015comprehensive}—we propose to apply a similar formulation to ``safely'' learning from imperfect demonstrations. Accordingly, we introduce \textbf{ConsMimic}, a CMDP‐based policy optimization framework that adaptively optimizes a stylistic imitation objective while rigorously keeping task performance above a user-set optimality threshold.

ConsMimic employs a multi-critic architecture~\citep{mysore2022multi} to independently estimate task-specific and style-specific values. These separate reward signals are adaptively combined using a self-adjustable Lagrangian multiplier, dynamically adjusting in response to observed violations of task optimality constraints~\citep{achiam2017constrained}. Additionally, we propose a symmetry-augmented style reward formulation to mitigate mode collapse that often plagues adversarial imitation learning, particularly when demonstrations are poorly aligned with task objectives. We validate our approach in three increasingly complex settings: (1) goal-reaching tasks on a Franka arm; (2) velocity tracking on a quadruped; and (3) velocity tracking over challenging terrains on a humanoid. In each scenario, demonstration data imperfectly match task conditions, yet still provide essential stylistic cues, such as agile quadruped gaits or coordinated arm-leg movements for humanoid locomotion. Our experiments demonstrate that ConsMimic allows robots to automatically determine when and how to utilize partial demonstrations, effectively preserving stylistic behaviors without compromising overall task performance. Supplementary videos for this work are available at \href{https://sites.google.com/view/consmimic}{our website.}

\textbf{In summary, our main contributions are:}

\begin{enumerate}
\item A CMDP-based policy optimization framework with a self-adjustable Lagrangian multiplier that explicitly enforces task optimality, enabling ``safe'' incorporation of stylistic cues from imperfect demonstrations.
\item A novel symmetry-augmented style reward formulation that effectively counteracts the mode collapse commonly induced by task-demo misalignments.
\item Comprehensive empirical validation on various robots through simulations and on ANYmal-D hardware, demonstrating effectiveness and generalization across various tasks and robotic platforms.
\end{enumerate}


\section{Related Work}
Deep reinforcement learning has brought great advances in the quadruped locomotion domain, where agents trained with parallel sampling~\citep{rudin2022learning} and domain randomization~\citep{tobin2017domain} are capable of transferring the learned policies from the simulation to the real world without finetuning. These policies often focus on optimizing task-specific objectives, such as speed or stability. However, such objectives alone can lead to unnatural, jerky movements that reduce robustness and realism.

To address these limitations, researchers have turned to more expressive reward designs that encode motion priors. By shaping the reward to discourage undesirable behaviors and guide agents toward plausible gait patterns, methods such as those in~\citep{lee2020learning, miki2022learning, shao2021learning} achieve more stable motion. Nevertheless, such priors are typically handcrafted and closely tied to the target task~\citep{margolis2023walk, yang2020multi, hu2024dexdribbler}, which limits their generalizability. A natural step forward is to move beyond manually crafted priors by leveraging demonstrations. Using retargeted motion capture data, researchers have trained policies to mimic skills exhibited by humans or animals. These skills can then be composed or reused in downstream tasks~\citep{bohez2022imitate, han2024lifelike}. In particular, adversarial imitation frameworks such as AMP~\citep{peng2021amp} and its successor ASE~\citep{peng2022ase} have proven effective in combining motion imitation with task execution, allowing smooth, agile behaviors in quadrupeds~\citep{escontrela2022adversarial, wu2023learning} and humanoids~\citep{tang2024humanmimic}. Further work~\citep{li2023learning} demonstrated that even highly dynamic motions, such as backflips, can be achieved with adversarial objectives, powered by frameworks like Wasserstein GANs~\citep{adler2018banach}.

Despite their success, these approaches implicitly assume that demonstration data will always help or at least not hinder task performance. In practice, this assumption does not hold: high-performing policies often require generalization to diverse terrains and command ranges~\citep{tobin2017domain}, while demonstrations may be limited, misaligned, or collected under drastically different conditions. Blindly following such data can negatively impact performance, especially when task requirements conflict with stylistic cues. Our work builds upon these insights and proposes a principled framework to reconcile this conflict. By adopting a CMDP-based policy optimization framework and using a self-adjustable Lagrangian multiplier to adjust imitation weight, we allow the agent to selectively learn from imperfect demonstrations without compromising its ability to learn robust task policies.

\section{Preliminary}

Crafting explicit reward functions for stylistic behaviors in robotics is often challenging due to complexity or sparsity, making direct imitation from expert demonstrations a practical alternative. One widely used technique is motion clip tracking, as introduced by \citet{peng2018deepmimic}. This method rewards the agent based on how closely its trajectory aligns with a provided reference motion clip at each timestep:
\begin{equation}
\label{eq:mimic}
r^s_{\text{track}} = \exp\left(-\sum_i w_i ( s_i - \hat{s}_i )^2 \right),
\end{equation}
where $s_i$ and $\hat{s}_i$ are subgroups of the agent and demonstration states, respectively, and $w_i$ are weighting coefficients.

Although effective in structured settings, tracking-based methods struggle when demonstrations are unstructured or not synchronized with tasks. To improve generalization, adversarial methods like Adversarial Motion Priors (AMP)~\citep{peng2021amp} employ a discriminator $D_\phi$ parameterized by $\phi$ trained to differentiate transitions from expert demonstrations and agent-generated transitions. The adversarial style reward is defined as:
\begin{equation}
\label{eq:amp}
r^s_{\text{adv}}(s_t, s_{t+1}) = \max\left(0, 1 - 0.25\left(D_\phi(\Phi(s_t), \Phi(s_{t+1})) - 1\right)^2\right),
\end{equation}
where $\Phi$ denotes a feature selector and the discriminator parameters $\phi$ are optimized by minimizing the loss:
\begin{equation}
\begin{aligned}
\label{eq:disc_loss}
    \arg\min_{\phi} \quad & \mathbb{E}_{(s,s')\sim d^M}\left[(D_\phi(\Phi(s), \Phi(s')) - 1)^2\right] 
    + \mathbb{E}_{(s,s')\sim d^{\pi}}\left[(D_\phi(\Phi(s), \Phi(s')) + 1)^2\right] \\
    &+ \frac{w_{\text{gp}}}{2}\mathbb{E}_{(s,s')\sim d^{M}}\left[
    \|\nabla_{\phi} D_\phi(\phi)\|^2\big|_{\phi=(\Phi(s), \Phi(s'))}
    \right],
\end{aligned}
\end{equation}
where $d^M$ and $d^\pi$ denote distributions of demonstration and policy-generated data, respectively. The last term is a gradient penalty weighted by $w_{\text{gp}}$ used to stabilize the training process. This adversarial framework promotes generalization to partial or imperfect demonstrations. In our work, we selectively apply tracking-based methods for manipulation tasks while adversarial imitation methods for locomotion tasks.

\section{Approach}
\begin{figure}[ht]
    \centering
    \includegraphics[width=1.0\linewidth]{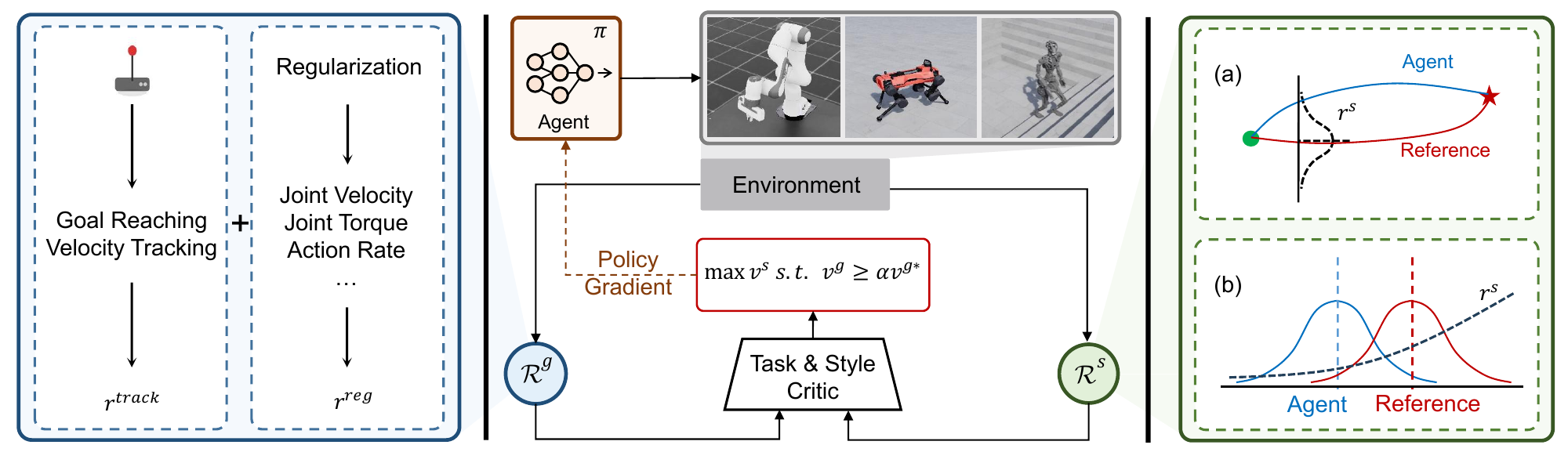}
    \caption{\textbf{ConsMimic Overview.} Given a reference dataset, ConsMimic calculates \textcolor[rgb]{0.0, 0.5, 0.0}{style reward} using either (a) motion clip tracking or (b) adversarial imitation learning methods. Such \textcolor[rgb]{0.0, 0.5, 0.0}{style reward ($R^s$)} 
    are combined with \textcolor[rgb]{0.0, 0.3, 0.6}{task reward ($R^g$)} within a constrained optimization framework illustrated \textcolor[rgb]{0.6, 0.0, 0.0}{in the red frame}. Separate critic networks estimate task and style advantages, which are subsequently weighted by a self-adjustable Lagrangian multiplier and finally used to optimize the policy.}
    \label{fig:consmimic}
\end{figure}

In this section, we present \textbf{ConsMimic}—a CMDP-based training pipeline for learning style-aware behaviors under task optimality constraints. An overview of the framework is illustrated in Fig.~\ref{fig:consmimic}, and the complete pseudocode is provided in Algorithm~\ref{alg:consmimic}.

\textbf{Constrained Style Learning under Task Optimality.}
In robotic tasks that require simultaneous task completion and stylistic imitation, effectively balancing these objectives is crucial but challenging. Typically, we categorize the rewards into two distinct groups: a manually designed demonstration-independent task reward group $r^g$, and a demonstration-driven style reward group $r^s$. Formally, we model this scenario as an extended Markov Decision Process (MDP):
$\mathcal{M} = \langle \mathcal{S}, \mathcal{A}, P, R^g, R^s, \mu, \gamma \rangle,$
where $\mathcal{S}$ is the state space, $\mathcal{A}$ denotes the action space, $P(s'|s,a)$ represents state transition probabilities, and reward functions $R^g, R^s: \mathcal{S} \times \mathcal{A} \times \mathcal{S} \rightarrow \mathbb{R}$ output scalar task and style rewards, respectively. $\mu$ defines the initial state distribution, and $\gamma \in (0,1)$ is the discount factor. The objective is to find a policy $\pi$ maximizing the expected cumulative discounted rewards from both reward groups:
\begin{equation}
\label{eq:v}
J_\pi = \mathbb{E}_{\pi}\left[\sum_{t=0}^{\infty} \gamma^t \left(r^g(s_t, a_t, s_{t+1}) + r^s(s_t, a_t, s_{t+1})\right)\right].
\end{equation}
Previous work typically employs hand-tuned weighted combinations of task and style rewards as the optimization objective. However, selecting appropriate weights can be time-consuming and provides no explicit guarantee of near-optimal task performance. To address this, we propose a CMDP-based policy optimization framework that guides the agent in determining \emph{when and how much} to extract stylistic cues from demonstration data while maintaining strong task performance. By explicitly incorporating a task optimality constraint, our approach ensures that the learned policy remains near-optimal. Formally, we define the CMDP as follows:
\begin{equation}
\label{eq:cmdp}
\max_{\theta}\; v^s(\pi_\theta) \quad \text{subject to} \quad v^g(\pi) \geq \alpha v^{g\star},
\end{equation}
where $ v^s(\pi_\theta) $ is the expected style value under policy $\pi_\theta$, $ v^g(\pi_\theta) $ is the expected task value, and $ v^{g\star} $ denotes the optimal achievable task performance. The parameter $\alpha \in [0,1]$ specifies the threshold for acceptable task performance relative to optimality.

In practice, we reduce the CMDP in Eq.~(\ref{eq:cmdp}) to a regular MDP using Lagrangian method~\citep{borkar2005actor, zahavy2022discovering}, resulting in the dual problem given by:
\begin{equation}
\label{eq:lagrangian_loss}
\min_{\lambda \geq 0} \max_{\theta} \mathcal{L}(\theta, \lambda) = v^s(\pi_\theta) + \lambda\left(v^g(\pi_\theta) - \alpha v^{g\star}\right),
\end{equation}
where $\lambda$ is the Lagrangian multiplier adaptively balancing task and style learning. To solve Eq. (\ref{eq:lagrangian_loss}), we alternatively optimize $\theta$ and $\lambda$. The optimization with respect to $\lambda$ can be expressed as 
\begin{equation}
    \min_{\lambda \geq 0} \lambda\left(v^g(\pi) - \alpha v^{g\star}\right).
\end{equation}
Intuitively, if task performance falls below the specified threshold, $\lambda$ increases, thereby emphasizing task objectives. In contrast, when task performance meets or exceeds the threshold, $\lambda$ decreases, allowing greater focus on style learning. Following previous work~\citep{zahavy2022discovering, cheng2024learning} that 
trades exact saddle-point optimality for empirical stability, we introduce a bounded multiplier via a sigmoid activation on normalized advantages to stabilize Proximal Policy Optimization (PPO)~\citep{schulman2017proximal} process, corresponding to the optimization of policy parameters $\theta$:
\begin{equation}
\label{eq:adv}
A = \sigma(\lambda)\tilde{A}^g + \left(1 - \sigma(\lambda)\right)\tilde{A}^s,
\end{equation}
where $\sigma(\cdot)$ denotes the sigmoid function, and $\tilde{A}^g$, $\tilde{A}^s$ represent normalized task and style advantages, respectively, computed via separate critic networks with Generalized Advantage Estimation (GAE)~\citep{schulman2015gae}. This bounded multiplier approach ensures stable training dynamics and adaptively balances the trade-off between task and stylistic objectives throughout the training process.

\textbf{Online Update of the Task Constraint.}
Although we can balance task and style objectives via a CMDP formulation, specifying an ``oracle'' optimal baseline $v^{g\star}$ beforehand is nontrivial. An overly optimistic oracle makes the constraint infeasible, while an underestimate yields suboptimal final performance. To avoid this, we introduce a \emph{warm-up} phase with the imitation weight set to zero, letting the policy optimize only the task reward; the statistical average task value of the converged warm-up policy seeds an initial $v_g^*$. During subsequent joint training, we monotonically update $v^{g\star}$ based on the best statistical task value $v^g(\pi)$ ever seen along the training process:
\begin{equation}
\label{eq:cons_update}
  v^{g\star} \;\leftarrow\;
  \max\Bigl(v^{g\star},\;v^g(\pi)\Bigr).
\end{equation}
Since $v^{g\star}$ reflects empirically demonstrated performance, constraint
$
  v^g(\pi)\;\geq\;\alpha\,v^{g\star}
$
remains feasible throughout the training. Moreover, once the policy exceeds the baseline, it must maintain at least a fraction $\alpha$ of the new best value of the task. This ensures near-optimal task performance is preserved when style imitation has positive influence over the task performance.

\textbf{Symmetric Augmented Style Learning.}
Symmetry is fundamental in robotic tasks, such as balanced locomotion with coordinated arm movements of humanoids, as it ensures harmonious and robust motion. However, adversarial imitation learning often struggles to capture these symmetric patterns due to the common issue of mode collapse in GAN training~\citep{adler2018banach}. This problem becomes particularly severe when the demonstration data are not well aligned with the task objectives, causing the discriminator to dominate, reducing the informativeness of its feedback. In our framework, this issue is primarily manifested by the policy repeatedly reproducing only a sub-segment of the periodic locomotion cycle, rather than capturing the full range of motion typically observed in natural locomotion patterns. The symmetry-augmentation method at policy optimization level proposed by~\citet{mittal2024symmetry} cannot effectively prevent this in this particular case due to the inherent asymmetry coming from the reward function. To address this, we inject symmetry directly into the reward. We define robot-specific symmetry transformation operators $L_g$, augmenting both demonstration and policy-generated data during training:
\begin{equation}
\mathcal{B}_{\text{sym}} = \mathcal{B} \cup \bigcup_{g \in G} L_g(\mathcal{B}),
\end{equation}
where $G$ is the set of symmetry transformations specific to the robot morphology, and $L_g(\mathcal{B}) = \{L_g(s, s') \mid (s, s') \in \mathcal{B}\}$. To reinforce symmetry in the learned policy explicitly, we compute the symmetry-augmented style reward by averaging discriminator outputs over all mirrored transitions:
\begin{equation}
\label{eq:sym_style}
r^s_{\text{sym}}(s_t, s_{t+1}) = \frac{1}{|G| + 1}\left[r^s_{\text{adv}}(s_t, s_{t+1}) + \sum_{g\in G} r^s_{\text{adv}}(L_g(s_t, s_{t+1}))\right].
\end{equation}
This generalized symmetry-aware formulation directly embeds symmetry constraints into the adversarial training objective, effectively counteracting biases resulting from partial, asymmetric demonstrations. By explicitly guiding the discriminator and the associated reward signal toward recognizing and enforcing symmetric behaviors, our method significantly enhances the robustness and efficiency of style generalization, leading to improved imitation quality even under challenging demonstration conditions.

\section{Experiments and Results}

\begin{figure}[ht]
    \centering
    \includegraphics[width=1.0\linewidth]{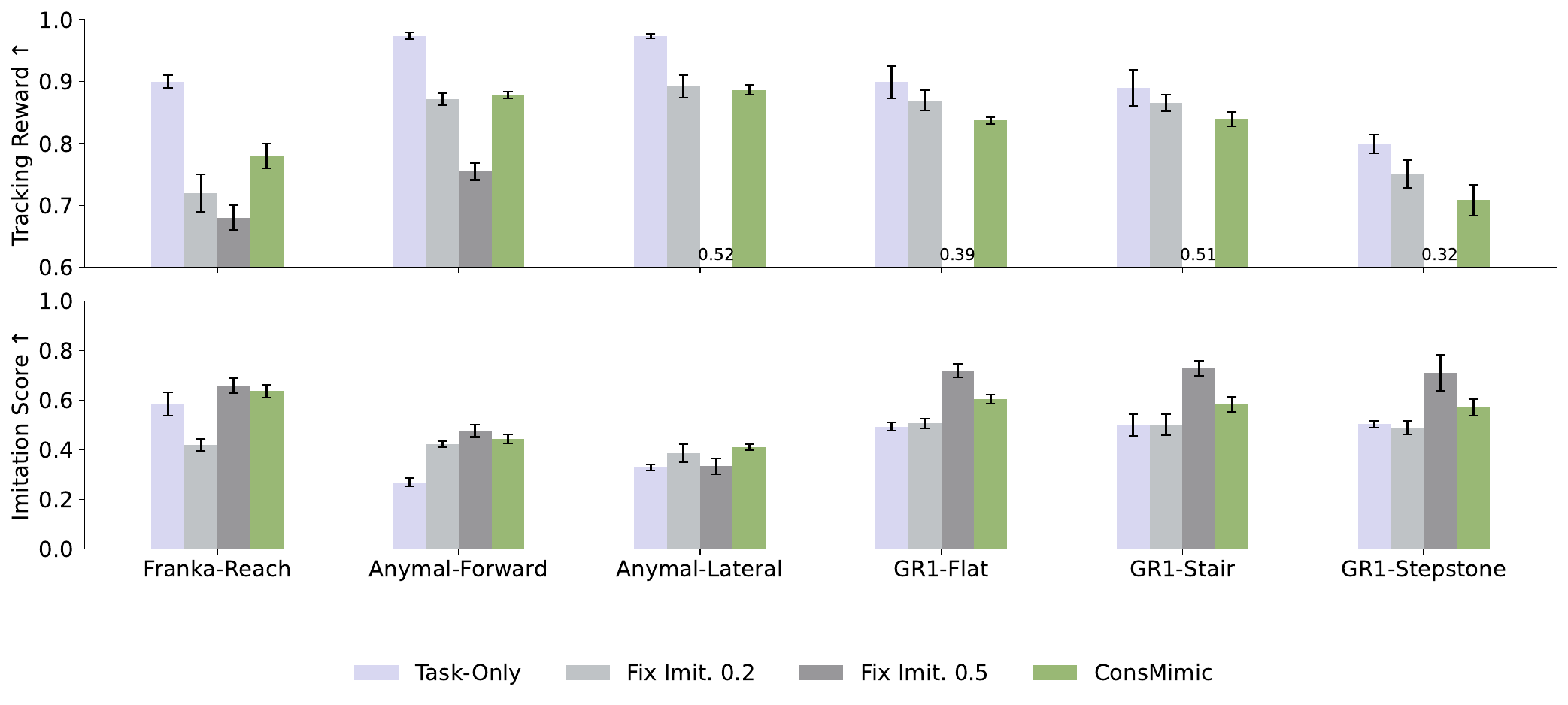}
    \caption{\textbf{Visualization Results across Tasks.} We report the mean and standard deviation over 5 seeds. The top row shows the tracking reward for each method. ConsMimic achieves task rewards comparable to the task-only baseline ($\omega^s_0$), demonstrating its ability to enforce near-optimal task performance. In contrast, the baseline with an aggressive imitation weight ($\omega^s_{0.5}$) struggles to learn how to complete the task. The bottom row presents the imitation scores. ConsMimic consistently outperforms all baselines that are capable of solving the tasks and only trails behind $\omega^s_{0.5}$, which achieves higher imitation at the cost of degraded task performance.}
    \label{fig:baseline}
\end{figure}

Our experiments aim to answer whether ConsMimic (i) achieves higher imitation quality while ensuring task optimality compared to task-only training and two fixed-weight baselines, (ii) improves motion symmetry using symmetry augmented style reward formulation, (iii) allows the task optimality threshold $\alpha$ to effectively control style--task trade-off, (iv) is beneficial to real-world applications.



\textbf{Q1. Style Learning Quality. } To rigorously validate ConsMimic, we designed comprehensive experiments across various robotic platforms and tasks that exhibit different degrees of misalignment between the demonstration tasks. Specifically, we evaluated performance on: (1) Franka-Reach, where the agent must reach a goal efficiently, while demonstrations follow stylistically sinusoidal trajectories; (2) Anymal-Forward and Anymal-Lateral tasks, where the ANYmal-D quadruped robot tracks forward or lateral velocities despite demonstrations consisting primarily of forward-trotting motions; and (3) GR1-Flat, GR1-Stair, and GR1-Stepstone, where a whole body humanoid robot GR1 is commanded to track velocities over diverse terrains while the reference motion is collected on flat ground. 

We compare ConsMimic ($\omega^s_\text{adapt}$) against two main baselines: (a) a task only baseline ($\omega^s_0$), which completely ignores stylistic imitation; and (b) fixed-weight baselines ($\omega^s_{0.5}$, $\omega^s_{0.2}$), that indicate that the style reward contribute 0.5 and 0.2 times the total reward group weight, respectively. We report detailed task formulations and task reward compositions in Appendix \ref{append:task}.

We report both achieved task performance and imitation quality across all six tasks in Fig.~\ref{fig:baseline}. For clarity and comparability, imitation scores are defined as:
\begin{equation}
S_{\text{imit}} = \max\{0, 1 - \text{DTW}(\tau^\pi, \tau^M)/\eta\},
\end{equation}
where DTW represents dynamic time warping~\cite{berndt1994using}, typically measuring distances between temporal sequences. $\tau^\pi$ and $\tau^M$ denote policy-generated and demonstration trajectories, respectively, and $\eta$ is a task-dependent normalization constant (e.g., $\eta = 20$ for manipulation, $\eta = 100$ for locomotion tasks). This formulation ensures that the scores are in the range $[0, 1]$.

As shown in Fig.~\ref{fig:baseline}, ConsMimic consistently achieves an effective balance between task performance and imitation quality. Specifically, in Franka-Reach, ConsMimic obtains a high task reward (0.78) and demonstrates significantly better imitation quality compared to the task-only baseline. In quadruped tasks, particularly Anymal-Lateral where style-task misalignment is prominent, ConsMimic achieves high task performance close to the task-only baseline while substantially surpassing fixed-weight baselines in imitation quality, highlighting its capacity to effectively leverage stylistic cues without compromising task objectives. For humanoid tasks, fixed-weight baselines either sacrifice task completion capability or experience a significant decline in stylistic imitation. For instance, $\omega^s_{0.5}$ demonstrates strong imitation but fails in complex scenarios like GR1-Stair and GR1-Stepstone, whereas $\omega^s_0$ maintains task performance but consistently scores lower in imitation quality. ConsMimic’s adaptive strategy consistently delivers better generalization by dynamically balancing style imitation and task requirements, enabling stable training and robust performance under realistic task-demo misalignments. We provide details for DTW calculation in Appendix \ref{append:eval}.

\begin{figure}[ht]
    \hspace{-0.2cm}
    \includegraphics[width=1.0\linewidth]{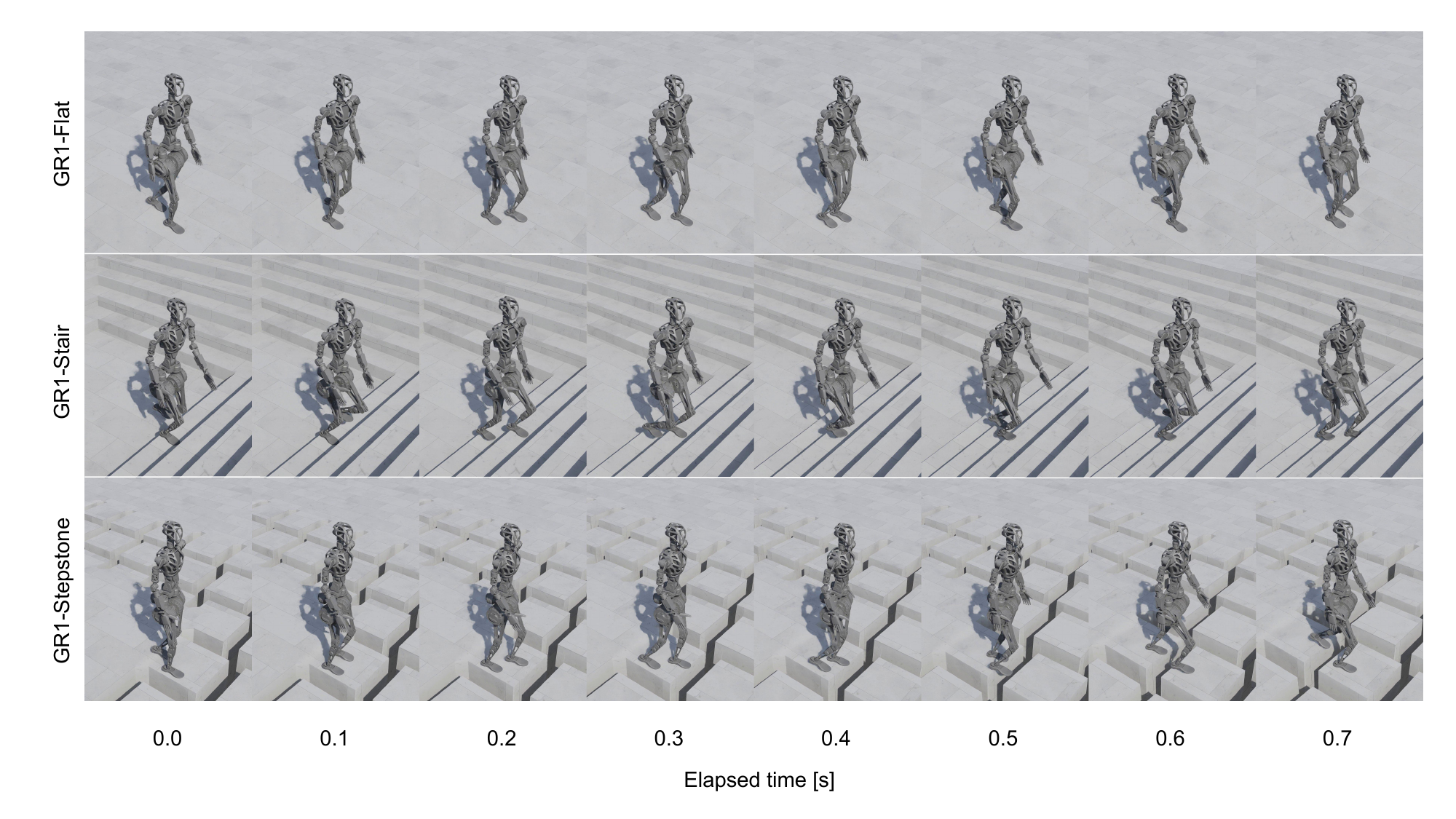}
    \caption{\textbf{Visualization Results of ConsMimic on GR1.} With symmetric augmented style learning, GR1 achieved symmetric and natural motion on both flat ground (in distribution) and stair \& stone ground (out of distribution).}
    \label{fig:gr1}
\end{figure}
\textbf{Q2. Symmetry-Augmented Style Learning. } 
To assess whether our symmetry-augmented reward improves motion symmetry, we measure a symmetry score calculated based on DTW distances between trajectories and their mirrored counterparts:
\begin{equation}
\label{eq:sym}
S_{\text{sym}} = \max\{0, 1 - \frac{1}{|G|} \sum_{g \in G} \frac{\text{DTW}(\tau^\pi, L_g(\tau^\pi))}{\eta}\},
\end{equation}
where $G$ is the set of predefined symmetry transformations, $L_g$ applies transformation $g$ to trajectory $\tau^\pi$, and $\eta$ is again a task-dependent normalization constant. 

\begin{figure}[ht]
    \centering
    \begin{minipage}[t]{0.46\textwidth}
        \centering
        \setlength{\tabcolsep}{7pt}
        \renewcommand{\arraystretch}{1.1}
        \setlength{\tabcolsep}{4pt}
        \vspace{-1.8cm}
            \begin{tabular}{lcc}
            \toprule
            Task & \makecell{Ours \\ \small(w/o sym aug)} & \makecell{Ours \\ \small(w/ sym aug)} \\
            \midrule
            GR1-Flat & $0.779_{\pm 0.021}$ & $0.814_{\pm 0.018}$ \\
            GR1-Stair & $0.741_{\pm 0.025}$ & $0.811_{\pm 0.020}$ \\
            GR1-Stepstone & $0.642_{\pm 0.030}$ & $0.722_{\pm 0.022}$ \\
            \bottomrule
        \end{tabular}
        \captionof{table}{%
        \textbf{Symmetry Analysis.} Symmetry scores $S_\text{sym}$ calculated by Eq. (\ref{eq:sym}) with $\eta = 100$ over 5 seeds. ConsMimic consistently improves symmetry by a large margin.}
        \label{tab:symmetry_score}
    \end{minipage}%
    \hfill
    \begin{minipage}[t]{0.50\textwidth}
        \centering
        \includegraphics[width=\linewidth]{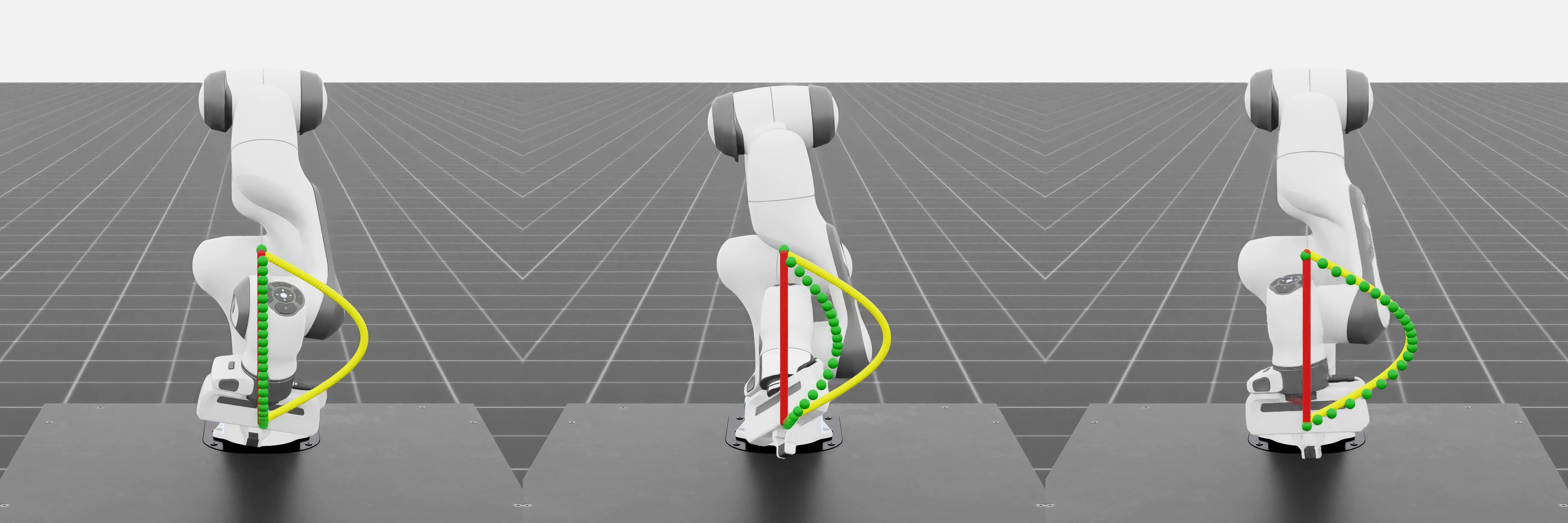}
        \caption{\textbf{Visualization of $\alpha$'s Effect on Franka-Reach.} Shown are trajectories for $\alpha = 1.0$ (left), $\alpha = 0.9$ (middle), and $\alpha = 0.8$ (right). The red line indicates the optimal task trajectory, the yellow line is the demonstration trajectory, and the green line shows our policy’s trajectory.}
        \label{fig:alpha}
    \end{minipage}
\end{figure}

As shown in Table~\ref{tab:symmetry_score}, incorporating our symmetry-augmented reward significantly improves the symmetry score across all GR1 locomotion tasks. This shows the effectiveness of our method on symmetric policy learning even from non-expert demonstrations. Our formulation enables the policy to generalize symmetric motion patterns across various terrains while preserving task optimality. We present visualization results of ConsMimic in GR1 tasks as Fig. \ref{fig:gr1}.

\textbf{Q3. Effectiveness of $\alpha$. } 
To assess whether $\alpha$ can effectively control the level of task optimality, we conducted experiments on the Franka-Reach task with $\alpha$ set to 0.8, 0.9, and 1.0. As shown in Fig.~\ref{fig:alpha}, when $\alpha = 1.0$, the policy completely disregards the demonstration's influence, strictly enforcing near-optimal task performance at the expense of stylistic cues. With $\alpha = 0.9$, the policy achieves high task performance while still incorporating essential stylistic features. When $\alpha = 0.8$, the agent fully assimilates the demonstration, yet still meets the required task constraints. These results demonstrate that $\alpha$ is an effective parameter for modulating the balance between task execution and style imitation.

\textbf{Q4. ConsMimic on Real-World Tasks. } We further validate our framework on the ANYmal-D quadruped. The policy, trained on the Anymal-Forward task, is deployed on the ANYmal-D hardware in a zero-shot manner. The robot is commanded to move forward at a speed of 2 m/s and then return at the same speed. It completes 8 rounds with identical distances.

\begin{figure}
    \centering
    \includegraphics[width=1.0\linewidth]{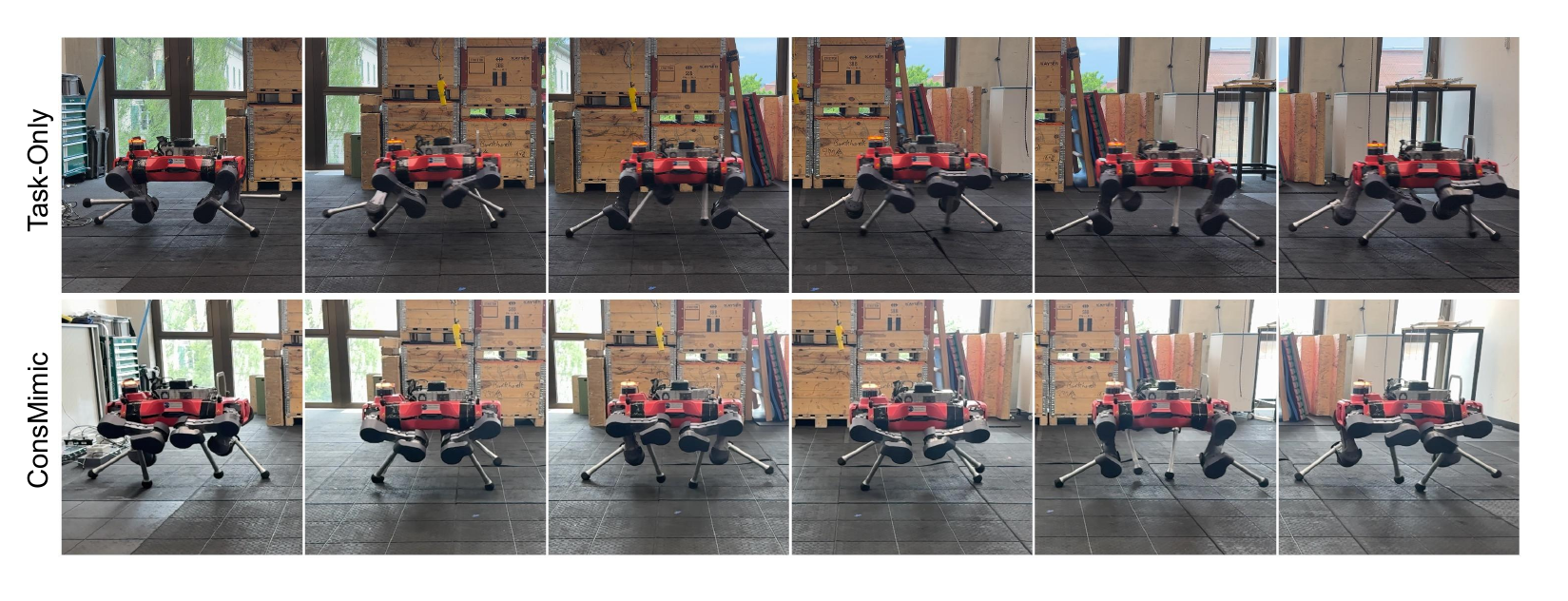}
    \caption{\textbf{Visualization of ANYmal-D's Locomotion in the Real World.} The top row show motions produced by policy trained by conventional task rewards while the bottom row trained by ConsMimic. ConsMimic enables the robot to achieve a more natural, agile trotting gait pattern.}
    \label{fig:anymal-d-locomotion}
\end{figure}

We analyze the motion styles produced by ConsMimic relative to the provided reference motion dataset. As illustrated in Figure~\ref{fig:anymal-d-locomotion}, policies trained within the ConsMimic framework achieve more natural and agile trotting gaits, a style difficult to learn solely from pure RL task rewards.

\begin{wraptable}{r}{0.45\linewidth}
    \centering
    \vspace{-0.3cm}
    \begin{tabular}{lcc}
    \toprule
    Metric & Task-Only & ConsMimic \\
    \midrule
    $W_\text{mech}$ (J)  & $1337 \pm 515$ & $1143 \pm 450$ \\
    $T_\text{air}$ (s)   & $0.28 \pm 0.02$ & $0.37 \pm 0.04$ \\
    \bottomrule
    \end{tabular}
    \caption{\textbf{Motion Analysis.} Policy trained with ConsMimic demonstrates lower energy usage and more dynamic motion.}
    \label{tab:real_world_power}
    \vspace{-1em}
\end{wraptable}

To quantitatively evaluate the quality of the motion, we measure two key metrics: the mechanical work ($W_\text{mech}$) done by the robot (computed as $\sum \tau \cdot \dot{\theta}$ per episode), and the average foot-air time (FAT, $T_\text{air}$) per step. As shown in Table~\ref{tab:real_world_power}, the ConsMimic-trained controller demonstrates superior energy efficiency and greater dynamism, characterized by lower mechanical energy consumption and increased foot-air time. The quantification results confirm that ConsMimic effectively converts stylistic motion imitation into tangible real-world performance enhancements, highlighting its practical applicability and robustness in real-world robotic locomotion tasks.

\section{Conclusion}
In this work, we introduced ConsMimic, a novel CMDP-based style learning framework designed to explicitly enforce task optimality while learning from imperfect demonstrations. Our method introduces a self-adjustable Lagrangian multiplier to automatically balance the trade-off between style learning and task learning and leverage symmetry augmented style reward formulation to extract symmetric patterns from motion reference. We validate our methods across robot platforms in simulator and real world. The experimental results show ConsMimic's potential as a practical and generalizable approach for real-world robotic style synthesizing tasks especially when expert demonstration is hard to access.

\section{Limitation}

Despite the promising results, our approach has limitations. Specifically, the framework does not explicitly discriminate between beneficial and detrimental features within demonstrations. Even when overall demonstrations are flawed or misaligned, there might still be valuable stylistic cues worth extracting. Future work should explore techniques to selectively identify and leverage beneficial demonstration features, potentially incorporating mechanisms such as attention or feature weighting to enhance the robustness and adaptability of imitation learning.

\clearpage
\acknowledgments{This research was supported by the ETH AI Center and the Swiss National Science Foundation through the National Centre of Competence in Automation (NCCR automation).}


\bibliography{example}  

\appendix

\section{Task Representation}
\label{append:task}
\subsection{Franka}

\textbf{Observation and Action Spaces.}  
The observation space for Franka consists of joint positions, joint velocities, the end-effector's target pose, and the policy’s previous actions. The action space comprises the target joint positions. The demonstration trajectory is represented by end-effector poses, as detailed in Table~\ref{tab:franka_spaces}.

\begin{table}[h]
\centering
\caption{Observation, action, and demonstration spaces for the Franka arm}
\label{tab:franka_spaces}
\begin{tabular}{llcc}
\toprule
\textbf{Category} & \textbf{Entry} & \textbf{Symbol} & \textbf{Dimension} \\
\midrule
\multirow{5}{*}{Observation} 
& Joint positions & $q$ & 7 \\
& Joint velocities & $\dot{q}$ & 7 \\
& End-effector position target & $p^*$ & 3 \\
& End-effector orientation target & $\theta^*$ & 4 \\
& Policy's last actions & $a_{t-1}$ & 7 \\
\midrule
Action & Target joint positions & $q^*$ & 7 \\
\midrule
Demonstration & End-effector poses & $x_t$ & 7 \\
\bottomrule
\end{tabular}
\end{table}

\textbf{Reward Formulation.}  
The task reward includes end-effector position and orientation tracking terms, action smoothness and regularization penalties, and a style imitation term. The style reward is defined as a demonstration trajectory tracking loss (see Eq.~\ref{eq:mimic}). Table~\ref{tab:franka_rewards} summarizes the weights for each term.

\begin{table}[h]
\centering
\caption{Task reward terms for the Franka arm}
\label{tab:franka_rewards}
\begin{tabular}{lcc}
\toprule
\textbf{Term} & \textbf{Function} & \textbf{Weight} \\
\midrule
EE tracking (coarse) & $||x - x^*||_2$ & $-0.5$ \\
EE tracking (fine-grained) & $1 - \tanh(||x - x^*||_2)$ & $1.0$ \\
EE orientation tracking & $||\theta - \theta^*||_{\text{quat}}$ & $-0.1$ \\
Action rate penalty & $||a_t - a_{t-1}||_2^2$ & $-0.01$ \\
Joint velocity penalty & $||\dot{q}||_2^2$ & $-0.01$ \\
\bottomrule
\end{tabular}
\end{table}

\textbf{Task Definition.}  
In the Franka-Reach task, the robot's end effector is commanded to reach a target pose where the position lies within $x \in [0.35, 0.45]$, $y \in [-0.05, 0.05]$, $z \in [0.20, 0.30]$ and the pitch angle satisfies $\theta_\text{pitch} \in [-\pi, \pi]$, all defined in the local frame.

\clearpage

\subsection{Anymal}

\textbf{Observation and Action Spaces.}  
The observation space for Anymal includes base velocities in the local frame, gravity projection, command inputs, joint positions, joint velocities, and past actions. The action space consists of target joint positions. The demonstration state space includes base velocities (local), gravity projection, joint positions, and joint velocities. These are detailed in Table~\ref{tab:anymal_spaces}.

\begin{table}[h]
\centering
\caption{Observation, action, and demonstration spaces for Anymal}
\label{tab:anymal_spaces}
\begin{tabular}{llcc}
\toprule
\textbf{Category} & \textbf{Entry} & \textbf{Symbol} & \textbf{Dimension} \\
\midrule
\multirow{7}{*}{Observation}
& Base linear velocity (local) & $v_\text{lin}^{\text{base}}$ & 3 \\
& Base angular velocity (local) & $v_\text{ang}^{\text{base}}$ & 3 \\
& Projected gravity & $g_\text{proj}$ & 3 \\
& Velocity commands & $v_\text{cmd}$ & 3 \\
& Joint positions & $q$ & 12 \\
& Joint velocities & $\dot{q}$ & 12 \\
& Previous actions & $a_{t-1}$ & 12 \\
\midrule
Action & Target joint positions & $q^*$ & 12 \\
\midrule
\multirow{4}{*}{Demonstration}
& Base linear velocity (local) & $v_\text{lin}^{\text{base}}$ & 3 \\
& Base angular velocity (local) & $v_\text{ang}^{\text{base}}$ & 3 \\
& Projected gravity & $g_\text{proj}$ & 3 \\
& Joint positions and velocities & $[q, \dot{q}]$ & 24 \\
\bottomrule
\end{tabular}
\end{table}

\textbf{Reward Formulation.}  
The task reward for Anymal includes tracking of commanded base velocities, as well as penalties on vertical motion, joint effort, energy consumption, flatness of base orientation, joint limit violations, and undesired contacts. The style reward is predicted according to Eq.~\ref{eq:sym_style}. All reward terms and their weights are listed in Table~\ref{tab:anymal_rewards}.

\begin{table}[h]
\centering
\caption{Reward terms for Anymal velocity tracking}
\label{tab:anymal_rewards}
\begin{tabular}{lcc}
\toprule
\textbf{Term} & \textbf{Function} & \textbf{Weight} \\
\midrule
Track linear velocity (xy) & $\exp\left(-\frac{||v_{xy} - v_{xy}^*||^2}{\sigma^2}\right)$ & $1.0$ \\
Track angular velocity (z) & $\exp\left(-\frac{||\omega_z - \omega_z^*||^2}{\sigma^2}\right)$ & $0.5$ \\
Vertical linear velocity penalty & $||v_z||_2^2$ & $-2.0$ \\
Angular velocity penalty (xy) & $||\omega_{xy}||_2^2$ & $-0.05$ \\
Joint torque penalty & $||\tau||_2^2$ & $-2.5\text{e}{-5}$ \\
Joint acceleration penalty & $||\ddot{q}||_2^2$ & $-2.5\text{e}{-7}$ \\
Action rate penalty & $||a_t - a_{t-1}||_2^2$ & $-0.01$ \\
Power consumption & $\sum \tau \cdot \dot{q}$ & $-5\text{e}{-5}$ \\
Feet air time reward & $\mathbb{1}_{t_\text{air} > 0.5}$ & $0.125$ \\
Undesired contacts (thigh) & $\mathbb{1}_{\text{contact}}$ & $-1.0$ \\
Flat orientation penalty & $||g_{b,xy}||_2^2$ & $-5.0$ \\
Joint limit violation penalty & $\sum_i \left[ \max(0, q_i - q_{\max,i}, q_{\min,i} - q_i) \right]$ & $-1.0$ \\
\bottomrule
\end{tabular}
\end{table}

\textbf{Task Definition.}  
In the Anymal-Forward task, the ANYmal robot is commanded to follow linear velocity commands in the body frame with $v_x \in [-3.0, 3.0]$ and angular velocity $\omega_z \in [-1.0, 1.0]$. In the Anymal-Lateral task, the robot is commanded to follow $v_y \in [-2.0, 2.0]$ and $\omega_z \in [-1.0, 1.0]$, also in the body frame.

\clearpage

\subsection{GR1}

\textbf{Observation and Action Spaces.}  
The observation space for GR1 includes base velocities in the local frame, projected gravity, command inputs, joint positions, joint velocities, past actions, and exteroceptive height scanning. The action space consists of target joint positions across legs, torso, shoulders, and elbows. The AMP demonstration state includes base motion, joint states, and foot positions. Details are summarized in Table~\ref{tab:gr1_spaces}.

\begin{table}[h]
\centering
\caption{Observation, action, and demonstration spaces for GR1}
\label{tab:gr1_spaces}
\begin{tabular}{llcc}
\toprule
\textbf{Category} & \textbf{Entry} & \textbf{Symbol} & \textbf{Dimension} \\
\midrule
\multirow{8}{*}{Observation}
& Base linear velocity & $v_\text{lin}^{\text{base}}$ & 3 \\
& Base angular velocity & $v_\text{ang}^{\text{base}}$ & 3 \\
& Projected gravity & $g_\text{proj}$ & 3 \\
& Velocity commands & $v_\text{cmd}$ & 3 \\
& Joint positions & $q_\text{rel}$ & 23 \\
& Joint velocities & $\dot{q}_\text{rel}$ & 23 \\
& Previous actions & $a_{t-1}$ & 23 \\
& Height scan & $h_\text{scan}$ & 173 \\
\midrule
Action & Target joint positions & $q^*$ & 23 \\
\midrule
\multirow{5}{*}{Demonstration}
& Base linear velocity & $v_\text{lin}^{\text{base}}$ & 3 \\
& Base angular velocity (local) & $v_\text{ang}^{\text{base}}$ & 3 \\
& Joint positions & $q$ & 23 \\
& Joint velocities & $\dot{q}$ & 23 \\
& Foot positions in local frame & $p_{\text{foot}}$ & 12 \\
\bottomrule
\end{tabular}
\end{table}

\textbf{Reward Formulation.}  
GR1's reward function combines task-level tracking, joint-level regularization, physical constraints, and biped-specific behavior shaping. It includes linear and angular velocity tracking, penalties on torque and joint deviation, foot-ground interaction shaping, and termination penalties. Table~\ref{tab:gr1_rewards} lists the main reward terms and weights.

\begin{table}[h]
\centering
\caption{Reward terms for GR1 rough terrain locomotion}
\label{tab:gr1_rewards}
\begin{tabular}{lcc}
\toprule
\textbf{Term} & \textbf{Function} & \textbf{Weight} \\
\midrule
Termination penalty & $\mathbb{1}_{\text{terminate}}$ & $-200.0$ \\
Track linear velocity (xy) & $\exp(-||v_{xy} - v_{xy}^*||^2 / \sigma^2)$ & $5.0$ \\
Track angular velocity (z) & $\exp(-||\omega_z - \omega_z^*||^2 / \sigma^2)$ & $3.0$ \\
Action rate (arms/legs) & $||a_t - a_{t-1}||_2^2$ & $-0.01$ \\
Action rate (2nd order) & $||a_t - 2a_{t-1} + a_{t-2}||_2^2$ & $-0.005$ \\
Joint torque penalty & $||\tau||_2^2$ & $-1\text{e}{-4}$ \\
Torque limit violation & $|\tau - \tau_\text{applied}|$ & $-0.002$ \\
Joint deviation penalty & $||q - q_\text{ref}||_2^2$ & $-0.5$ \\
Feet air time reward & $\mathbb{1}_{t_\text{air} > 0.4}$ & $1.0$ \\
Zero action (ankle roll) & $\mathbb{1}_{|a| > \epsilon} \cdot a^2$ & $-0.5$ \\
Joint limit violation & $\mathbb{1}_{q \notin [q_{\min}, q_{\max}]}$ & $-10.0$ \\
Power consumption & $\sum \tau \cdot \dot{q}$ & $-5\text{e}{-6}$ \\
Base angular velocity (xy) & $||\omega_{xy}||_2^2$ & $-0.05$ \\
Feet slide penalty & $||v_\text{slip}||$ when in contact & $-1.0$ \\
No-fly penalty & $\mathbb{1}_{\text{both feet airborne}}$ & $-5.0$ \\
Pelvis orientation & $||g_{\text{pelvis},xy}||_2^2$ & $-5.0$ \\
Torso orientation & $||g_{\text{torso},xy}||_2^2$ & $-5.0$ \\
\bottomrule
\end{tabular}
\end{table}

\textbf{Task Definition.}  
In all GR1 task, the GR1 robot is commanded to follow linear velocity commands in the body frame with $v_x \in [0.5, 2.0]$.
\clearpage

\section{Training Details}

\subsection{Training Pipeline}
We detail our training pipeline in the following algorithm:

\begin{algorithm}[ht]
    \caption{ConsMimic Training Pipeline}
    \label{alg:consmimic}
    \begin{algorithmic}[1]
        \STATE \textbf{Require:} Policy $\pi$, task critic $v^g$, style critic $v^s$, discriminator $D_\phi$, Lagrange multiplier $\lambda$, demonstrations $\mathcal{D}$, symmetry mappings $G$, threshold coefficient $\alpha$, Learning iterations $N$, Constraint update intervals $I_c$
        \STATE Initialize networks and rollout buffer $\mathcal{B}$
        \STATE Set optimal task value $v^g$ as the initial guess $v^{g*}$
        \FOR{learning iteration $i = 1, 2, \cdots, N$}
            \FOR{time step $t = 1, 2, \dots, T$}
                \STATE Collect transition $(s_t, a_t, s_{t+1}, r^g_t)$ using current policy $\pi$
                \STATE Compute symmetry-augmented style reward $r^s_{\text{sym},t}$ using Eq.~(\ref{eq:sym_style})
                \STATE Store $(s_t, a_t, s_{t+1}, r^g_t, r^s_{\text{sym},t})$ in rollout buffer $\mathcal{B}$
            \ENDFOR
            \STATE Compute TD targets for value updates
            \STATE Compute task advantage $A^g$ and style advantage $A^s$ using GAE
            \STATE Compute combined advantage using Eq.~(\ref{eq:adv}) ($\sigma(\lambda)$ is set to 1 during \emph{warmup} phase)
            \FOR{learning epoch $= 1, 2, \dots, K$}
                \STATE Sample mini-batches $b\sim\mathcal{B}$ 
                \STATE Update policy $\pi$, task critic $v^g$, and style critic $v^s$ using PPO
                \STATE Update discriminator $D_\phi$ using symmetry-augmented mini-batch $b_{\text{sym}}$ via Eq.~(\ref{eq:disc_loss})
                \STATE Update Lagrange multiplier $\lambda$ using Eq.~(\ref{eq:lagrangian_loss})

            \IF{$i \bmod I_c = 0$}
            \STATE Update constraint using Eq. \ref{eq:cons_update}
            \ENDIF
                
            \ENDFOR
        \ENDFOR
    \end{algorithmic}
\end{algorithm}

\subsection{Network Architecture}
The policy network, value network, discriminator network all consist of MLP layers, which is detailed in Table \ref{tab:network_architecture}

\begin{table}[h]
\centering
\caption{Network architectures used for each task}
\label{tab:network_architecture}
\begin{tabular}{lccc}
\toprule
\textbf{Task} & \textbf{Policy} & \textbf{Value} & \textbf{Discriminator} \\
\midrule
Franka & [64, 64] & [64, 64] & – \\
Anymal & [512, 256, 128] & [512, 256, 128] & [1024, 512] \\
GR1 & [512, 256, 128] & [512, 256, 128] & [1024, 512] \\
\bottomrule
\end{tabular}
\end{table}

\subsection{Training Parameters}

\begin{table}[ht]
\centering
\caption{ConsMimic training parameters.}
\begin{tabular}{llll}
\toprule
\textbf{Parameter} & \textbf{Value} & \textbf{Parameter} & \textbf{Value} \\
\midrule
Num Steps per Environment & 24 & Training Iterations & 20000 \\
clip range & 0.2 & entropy coef & 0.005 \\
mini batches & 4 & learning rate & 1e-3 \\
discount factor & 0.99 & $\alpha$ for Franka & 0.9 \\
$\alpha$ for Anymal & 0.7 & $\alpha$ for GR1 & 0.9 \\
\bottomrule
\end{tabular}
\end{table}

\clearpage
\section{Evaluation Details}
\label{append:eval}

\subsection{Terrain for Evaluation}

We evaluate GR1 locomotion performance under two challenging terrain settings: \textit{Stairs} and \textit{Stepping Stones}, each defined using custom terrain generator configurations in Isaac Lab. Key parameters and sub-terrain definitions are summarized below.

\textbf{Stairs.} The stair terrain consists of pyramid-style inverted stairs with varying step heights. The step height ranges from 0.05 to 0.27 meters; step width ranges from 0.30 to 0.40 meters.

\textbf{Stepping Stones.} This terrain contains high-frequency stepping-stone terrain with variable widths and distances. Stone height is up to 0.01 meters, width ranges from 0.55 to 1.0 meters; distance ranges from 0.1 to 0.2 meters.

\subsection{Imitation \& Symmetry Score Calculation}

We use dynamic time warping (DTW) to evaluate the distance between the policy-generated trajectory and the demonstration trajectory. The standard DTW implementation requires alignment of the start and end of two trajectories, which is impratical in our settings. We relax this constraint in our implementation as specified in Algorithm \ref{alg:dtw}

\begin{algorithm}[h]
\caption{Relaxed DTW Distance}
\label{alg:dtw}
\begin{algorithmic}[1]
\REQUIRE Two sequences: $\text{seq}_1 \in \mathbb{R}^{n \times d}$ and $\text{seq}_2 \in \mathbb{R}^{m \times d}$
\STATE Initialize DTW matrix: $D \in \mathbb{R}^{(n+1) \times (m+1)}$ with $D[0, :] \gets 0$, $D[:, 0] \gets \infty$
\FOR{$i = 1$ to $n$}
    \FOR{$j = 1$ to $m$}
        \STATE $c \gets \| \text{seq}_1[i] - \text{seq}_2[j] \|_2$
        \STATE $D[i, j] \gets c + \min\{ D[i-1, j], D[i, j-1], D[i-1, j-1] \}$
    \ENDFOR
\ENDFOR
\STATE \textbf{return} $\min_j D[n, j]$ \COMMENT{Relax end-alignment by taking minimal cost across final row}
\end{algorithmic}
\end{algorithm}

Note that we use end effector trajectory for Franka task while joint position trajectory for Anymal and GR1 tasks.

\end{document}